\documentclass{article}
\usepackage{spconf,amsmath,epsfig}

\usepackage{multirow}
\usepackage{makecell}
\usepackage{booktabs}
\usepackage{amssymb}
\usepackage{url}
\usepackage{bbding}
\usepackage[ruled,linesnumbered]{algorithm2e}

\let\OLDthebibliography\thebibliography
\renewcommand\thebibliography[1]{
  \OLDthebibliography{#1}
  \setlength{\parskip}{0pt}
  \setlength{\itemsep}{0pt plus 0.3ex}
}

\begin{document}\sloppy

\def\x{{\mathbf x}}
\def\L{{\cal L}}

\title{SSyncOA: Self-synchronizing Object-aligned Watermarking \\ to Resist Cropping-paste Attacks}
%
\name{Chengxin Zhao\textsuperscript{1}, Hefei Ling\textsuperscript{1*}, Sijing Xie\textsuperscript{1}, Han Fang\textsuperscript{2}, Yaokun Fang\textsuperscript{1}, Nan Sun\textsuperscript{1}}
\address{\textsuperscript{1}Huazhong University of Science and Technology, China \\ \textsuperscript{2}National University of Singapore, Singapore}

\maketitle

\begin{abstract}

	Modern image processing tools have made it easy for attackers to crop the region or object of interest in images and paste it into other images. The challenge this cropping-paste attack poses to the watermarking technology is that it breaks the synchronization of the image watermark, introducing multiple superimposed desynchronization distortions, such as rotation, scaling, and translation. However, current watermarking methods can only resist a single type of desynchronization and cannot be applied to protect the object's copyright under the cropping-paste attack. With the finding that the key to resisting the cropping-paste attack lies in robust features of the object to protect, this paper proposes a self-synchronizing object-aligned watermarking method, called SSyncOA. Specifically, we first constrain the watermarked region to be aligned with the protected object, and then synchronize the watermark's translation, rotation, and scaling distortions by normalizing the object invariant features, i.e., its centroid, principal orientation, and minimum bounding square, respectively.  To make the watermark embedded in the protected object, we introduce the object-aligned watermarking model, which incorporates the real cropping-paste attack into the encoder-noise layer-decoder pipeline and is optimized end-to-end. Besides, we illustrate the effect of different desynchronization distortions on the watermark training, which confirms the necessity of the self-synchronization process. Extensive experiments demonstrate the superiority of our method over other SOTAs.

\end{abstract}
\section{Introduction}
\label{sec:intro}

The advancement of image processing techniques has made the process of image editing convenient and user-friendly, but such a convenience on the other hand created new demands on copyright protection techniques, i.e., digital watermarking. In practical applications, in addition to manipulating entire images, regional editing of specific objects within images is also widespread. With the growing popularity of new digital assets, 
achieving finer-grained certification at the object level have become increasingly vital. Current watermarking methods typically apply watermarks to the entire image. However, object-based theft, namely cropping-paste attacks, would destroy the synchronization of image watermarks, leading to unsuccessful copyright certification. The cropping-paste attack commonly introduces multiple desynchronization distortions, including cropping, translation, rotation, and scaling. This highlights the need to address synchronization challenges in watermarking methods to effectively resist such attack.

\begin{figure}[t]
	\includegraphics[width=0.98\linewidth]{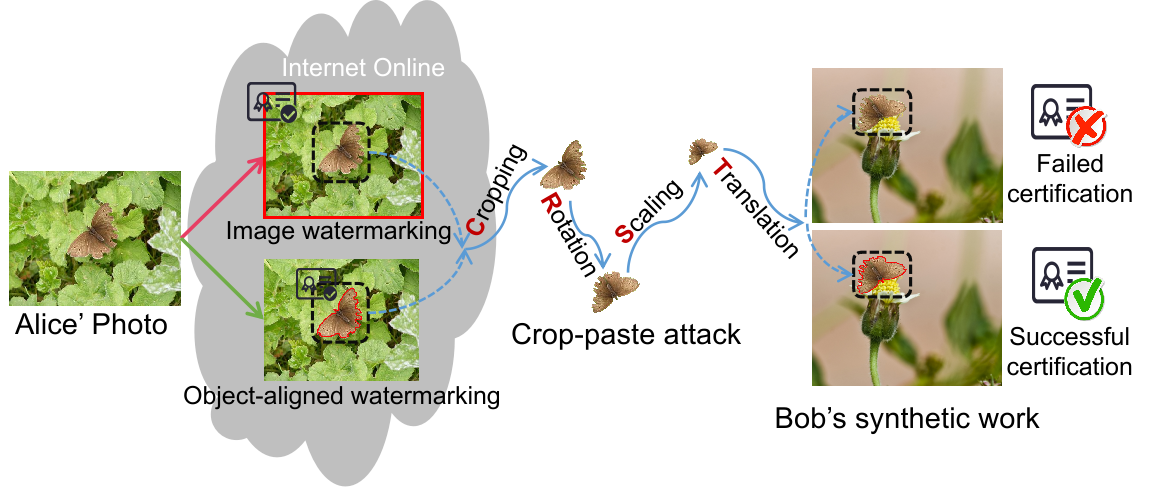}
	\centering
	\caption{Example of the cropping-paste attack, where the butterfly captured by Alice is stolen by Bob. The red outlines indicate the embedded region of the watermark. Due to various desynchronization distortions, the image watermark embedded in the photo is destroyed, resulting in failed certification. Here, we propose the watermarking scheme SSyncOA, which embeds the watermark in the object region and synchronizes the distortion through the object invariant features.}
	\label{fig:image_object}
\end{figure}

To resist desynchronization distortions, early works improve robustness through training. HiDDeN \cite{conf/eccv/ZhuKJF18} first adds the cropping distortion between the watermark encoding and decoding process, and achieves anti-cropping capability through end-to-end training.  SSL \cite{conf/icassp/FernandezSFJD22} proposes to embed watermarks in self-supervised latent spaces, and they improve the robustness against rotation and scaling by data augmentation.  Although these training-based methods provide a straightforward way to improve robustness, various superimposed desynchronization distortions would place a heavy burden on model optimization, resulting in severe visual quality degradation. Recently, synchronization-based methods have been proposed to address this problem. To synchronize the geometric distortion caused by camera shooting, StegaStamp \cite{conf/cvpr/TancikMN20} and Invisible Markers \cite{conf/cvpr/JiaGZMZY22} use location models to detect the four vertices of the embedded region in photos, and then synchronize the distortion by perspective transformation. However, they cannot resist cropping attacks. The watermarking robustness against multiple superimposed desynchronization distortions needs further improvement.




Although the cropping-paste attack poses a desynchronization challenge for image watermarking, it also introduces geometrically robust features in terms of the protected (attacked) object. As shown in Fig.\ref{fig:image_object}, the cropping-paste attack can be decomposed into object-based cropping, translation, rotation, and scaling. This motivates us to use the object invariant features to achieve watermark synchronization, given that these distortions do not alter the object region itself.

Based on this, we propose a self-synchronizing object-aligned watermarking scheme to resist the cropping-paste attack. We call the scheme SSyncOA, which consists of two main components: the self-synchronization process (SSync) and the object-aligned watermarking model (OA). For SSync, it first aligns the watermark region with the protected object region, then normalizes the region's centroid, principal orientation, and minimum bounding square to a default state. 
For OA, we apply SSync on both the encoder and decoder inputs so that the watermark region is geometrically synchronized. To ensure that OA embeds and extracts the watermark within the synchronized (object) region, we introduce the cropping-paste attack into the noise layer and optimize our model end-to-end. Through a joint optimization of SSync and OA, SSyncOA can successfully decode the watermark from the synthetic image subjected to a cropping-paste attack. The main contributions of this work are summarized as follows.

\begin{itemize}
	\item We propose a self-synchronizing object-aligned watermarking scheme that provides an approach for object-level copyright protection and significantly improves the visual quality of watermarked images.
	\item We introduce a segmentation-based watermarking region detection method that enables automatic blind watermark synchronization and extraction.
	\item We analyze the impact of desynchronization distortions on the training of the watermarking model.  Extensive comparisons with other SOTAs confirm the superiority of our method.
\end{itemize}

\begin{figure*}[t]
	\includegraphics[width=0.9\textwidth]{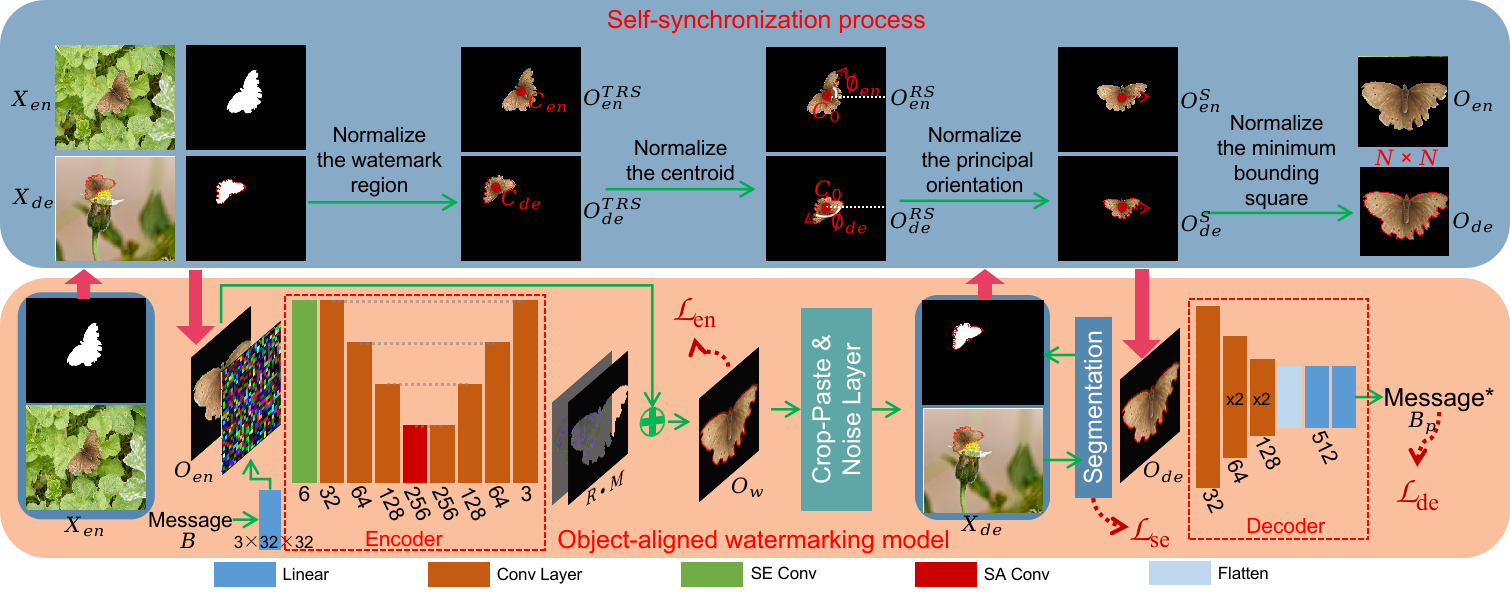}
	\centering
	\caption{Training pipeline of SSyncOA, which consists of the self-synchronization process (SSync) and the object-aligned watermarking model. Given the object image $X_{en}$ to be protected, it is first synchronized by SSync, then the synchronization result $O_{en}$ is fed to the encoder to generate the watermarked object $O_w$. To simulate the cropping-paste attack, $O_w$ is pasted into another background image and further distorted by the noise layer. Given the synthetic image $X_{de}$ to be authenticated, it is also first synchronized by SSync. The decoder takes the synchronized object $O_{de}$ as input and extracts the embedded message.}
	\label{fig:pipeline}
\end{figure*}

\section{Related Works}
\label{sec:related_works}

Existing desynchronization-distortion resilient watermarking methods can be roughly divided into two categories: invariant feature-based methods and synchronization-based methods. 

\textbf{Invariant feature-based watermarking methods.} Traditional invariant feature-based methods \cite{journals/tifs/KangHZ10, journals/tip/ZhangSCZWZZL11} embed watermarks in the geometrically invariant frequency domain to resist distortions. After HiDDeN \cite{conf/eccv/ZhuKJF18} proposed the Encoder-Noise layer-Decoder (END) watermarking architecture, most current watermarking schemes \cite{conf/mm/JiaFZ21, journals/tim/HuangLLYXC23, journals/tcsvt/Wang23} introduce the desynchronization distortions directly into the noise layer to improve robustness. However, for superimposed distortions, e.g., the cropping-paste attack, this simple strategy leads to unstable training and severe visual quality degradation. 

\textbf{Synchronization-based watermarking methods.} Traditional synchronization-based methods \cite{journals/tcsv/FangCHZMZY21} resort to embedding and matching synchronization templates, while recent works \cite{conf/cvpr/TancikMN20, conf/cvpr/JiaGZMZY22, conf/mm/GuoZLGZSL23} propose to use deep learning-based models to detect the watermark regions and then correct them.  However, most of them focus on the synchronization of the entire image and cannot be applied to resist the object-based cropping-paste attack.

\textbf{Object watermarking.} 
Several prior works \cite{journals/pr/GuoZS03, journals/prl/HoW04, conf/icip/PhamMYA07, journals/mta/GajPS16} have explored object watermarking, primarily relying on traditional methods with manual design. To the best of our knowledge, we are the first to combine object watermarking with deep learning,  achieving fully automatic localization, synchronization, and decoding. Additionally, our method attains significantly higher robustness, capacity, and visual quality compared to these earlier approaches.
\section{Method}
\label{sec:method}

As shown in Fig.\ref{fig:image_object}, during Bob's crop and paste attack, the watermark is first changed in shape due to cropping, and then is transformed due to superimposed rotation, scaling, and translation distortions. In direct decoding of the synthetic image, the geometry of the watermark region in the decoder inputs is completely inconsistent with the original embedding state, resulting in desynchronization and decoding failure.

To synchronize the watermark state for successful certification, we propose a self-synchronizing object-aligned watermarking method, called SSyncOA. Fig.\ref{fig:pipeline} presents the training pipeline, and we describe its two main components and the training loss below.

\subsection{Self-synchronization process}


The self-synchronization process is used to synchronize the watermark geometry between the image $X_{en}$ to be encoded and the image $X_{de}$ to be decoded. Here, we sequentially synchronize the four distortions by normalizing the corresponding invariant features to a default state. 


\textbf{Normalize the watermark region to synchronize cropping.} 
We first normalize the watermark region to be aligned with the protected (attacked) object region. It ensures the integrity of watermark information after the cropping attack. We achieve this by removing the background pixels in $X_{en}$ and $X_{de}$, resulting in $O_{en}^{TRS}$ and $O_{de}^{TRS}$. 
The cropping distortion is thereby synchronized by detecting the specific object before encoding and decoding.

\textbf{Normalize the centroid to synchronize translation.} The translation synchronization is to make the position of the watermark (object) region in $O_{de}^{TRS}$ is coincide with that in $O_{en}^{TRS}$. Given that the centroid, i.e., the gravity center of the object, remains invariant if its shape is unchanged, we normalize the centroid to the center of the inputs after the cropping synchronization. Here, we get the centroids $(x, y)$ using OpenCV, i.e., 
\begin{equation}
	(x, y) = (\frac{m_{10}}{m_{00}}, \frac{m_{01}}{m_{00}})
\end{equation}
where $m_{ij}$ is the object contour’s spatial moments. It is calculated by cv2.moments() \footnote{https://docs.opencv.org/4.8.0/d8/d23/classcv\_1\_1Moments.html}.
As shown in Fig.\ref{fig:pipeline}, by moving both the centroid $C_{en}$ of $O_{en}^{TRS}$ and the centroid $C_{de}$ of $O_{de}^{TRS}$ to the center of inputs $C_{0}$, the translation is synchronized, obtaining  $O_{de}^{RS}$ and $O_{de}^{RS}$.



\textbf{Normalize the principal orientation to synchronize rotation.} Rotation synchronization aims to makes the object principal orientation in $O_{de}^{RS}$ is the same as that in $O_{en}^{RS}$. When the centroid is fixed, the principal orientation is changed synchronously with the rotation. Here, the principal orientation $\phi$ is normalized to $0^\circ$ by rotating the object around its centroid, and it is calculated as 
\begin{equation}
	\phi = 0.5 \times \operatorname{arctan2}(\frac{2mu_{11}}{mu_{00}}, \frac{mu_{20}-mu_{02}}{mu_{00}})
\end{equation}
where $mu_{ij}$ is the object contour’s central moments. By rotating the $O_{en}^{RS}$ with $\phi_{en}$ degree and rotating $O_{de}^{RS}$ with $\phi_{de}$ degree, we get $O_{en}^{S}$ and $O_{en}^{S}$ that are rotation synchronized.


\textbf{Normalize the minimum bounding square to synchronize scaling.} Scaling synchronization is to make the scale of the object in $O_{de}^{S}$ is the same as that in $O_{en}^{S}$. Based on the above synchronization, we re-scale the minimum bounding square (MBS) of objects to $N \times N$ to achieve normalization. The MBS is obtained by padding the minimum bounding rectangle. By re-scaling the MBS of both $O_{en}^{S}$ and $O_{de}^{S}$, we finally get the geometry synchronized object (i.e., the watermark region) $O_{en}$ to be encoded and object $O_{de}$ to be decoded.

\subsection{Object-aligned watermarking model}


The object-aligned watermarking model is responsible for embedding and extracting watermarks from the specific object region. The model details are described below.


\textbf{Encoder.} 
We take the synchronized object $O_{en}$ as the host image, where the background pixels are zeroed except for the object to be protected. This ensures  that the encoder can only extract features from the object region. As for the watermark message, i.e., a 0/1 bit string, we perform a linear transform before concatenating it with $O_{en}$, which helps to increase message redundancy and spatially align the message with the object. The encoder outputs a residual map $R$, we remove the background by the object mask $M$, and then superimpose it onto $O_{en}$ to get the watermarked object $O_{w}$, i.e., $O_w = O_{en} + R \times M$.

\textbf{Copy-paste attack and the noise layer.}
Although the self-synchronization module allows the encoding and decoding process to be performed on the synchronized object images, there is no perfect synchronization in practice due to the cropping or sampling bias. Therefore, we include both the cropping-paste attack and the self-synchronization process in our end-to-end training process. Specifically, we randomly \textbf{R}otate($\le 45^\circ$), \textbf{S}cale($\in[0.75, 1.5]$), and paste (\textbf{T}ranslate) the encoded object $O_w$ onto other background images. The composed image is further distorted with randomly selected noise layers, including Gaussian Blur($\sigma_1 \le 2.$), Gaussian Noise($\sigma_2 \le 0.05$), and Pseudo-JPEG \cite{conf/cvpr/TancikMN20}(QF$\in$\{50,75\}, which is differentiable).



\textbf{Segmentation model and decoder.} Given the image to be decoded, i.e., $X_{de}$, the message extraction is automatic and blind: we first segment the watermark (object) region, then synchronize it based on the segmentation mask, and finally extract the message from the synchronized image. A simple Resnet-18 based UNet is used here as the segmentation model, and the decoder consists of several convolutional and linear layers as shown in Fig.\ref{fig:pipeline}.


\begin{table}[t]
	\centering
	\caption{Performance of the models trained with different noise layers.}
	\resizebox{\linewidth}{!}{
		\begin{tabular}{c|ccccc|c} \hline
			& None & R.S.T. & G-Blur & G-Noise & P-JPEG & Combined \\ \hline
			PSNR & 53.08 & 52.82 & 51.65 & 44.80 & 43.90 & 43.93 \\
			SSIM & 99.87 & 99.87 & 99.82 & 99.18 & 99.58 & 99.56 \\
			BAR & 99.95 & 99.20 & 97.74 & 97.57 & 96.89 & 98.08 \\ \hline
			IoU & 99.43 & 99.21 & 94.60 & 98.69 & 96.11 & 97.92  \\
			BAR$_{gt}$ & 99.95 & 99.20 & 98.78 & 98.00 & 97.04 & 98.68 \\
			\hline
		\end{tabular}
	}
	\label{tab:result}
\end{table}

\subsection{Training Loss}
There are three loss items to supervise our model training, including encoding loss, segmentation loss and decoding loss. \textbf{The encoding loss $\mathcal{L}_{en}$} is used to supervise the visual variation caused by encoding. We use the L2 norm of the residual map $R$ to constrain the pixel modification, and the LPIPS loss \cite{conf/cvpr/ZhangIESW18} is also adopted here to optimize visual perception. \textbf{The segmentation loss $\mathcal{L}_{se}$} is the Lovász hinge loss \cite{conf/cvpr/BermanTB18}, which computes a surrogate binary intersection-over-union between the predicted mask $M_{p}$ and its ground truth $M$. Since the decoding result is a 0/1 bit string, we directly use the binary cross entropy (BCE) between the predict message $B_p$ and the ground truth $B$ as \textbf{the decoding loss $\mathcal{L}_{de}$}. The total loss is the sum of them, i.e, 

\begin{gather}
	\mathcal{L}_{en} = \lambda_1 \|R\|_2 + \lambda_2 \operatorname{LPIPS}(O_{en}+R, O_{en}) \\
	\mathcal{L}_{se} = 1 - \frac{|M_{p} \cap M|}{|M_{p} \cup M|} \\
	\mathcal{L}_{de} = -\lambda_3[B \cdot \operatorname{log}B_p + (1-B) \cdot \operatorname{log}(1-B_p)]  \\
	\mathcal{L} = \mathcal{L}_{en} + \mathcal{L}_{se} + \mathcal{L}_{de}
\end{gather}
where $\lambda_1$, $\lambda_2$, $\lambda_3$ are weight factors, we set them to 1.5, 1.2, and 2 by default.

%
%

\section{Experiments}
\label{sec:experiments}
The salient object detection dataset DUTS \cite{conf/cvpr/WangLWF0YR17} is used here for providing object images and the corresponding masks. 
To align with the common capacity setting \cite{conf/eccv/ZhuKJF18} that embeds 30 bits into the 128$\times$128 image, we resize all images to 256 $\times$ 256 and filter out the images that the object occupies less than $\frac{1}{4}$. We end up with 8236 objects with various shapes for training, 1198 objects for validation, and 1198 objects for testing. Background images used in the cropping-paste attack are randomly selected from the original DUTS dataset. We resized them to 512 $\times$ 512. During training, we use Adam with a learning rate of 1e-4 to optimize parameters, the weight decay is set to 1e-5. The batch size is 12 and we train the model for 360,000 steps in total.
Bit Accuracy Rate (BAR) is used to measure the robustness of watermarking models. The PSNR(dB) and SSIM are used to evaluate the visual quality of the watermarked image. We use the IoU(\%) metric to indicate the segmentation performance.

\begin{figure}[t]
	\includegraphics[width=\linewidth]{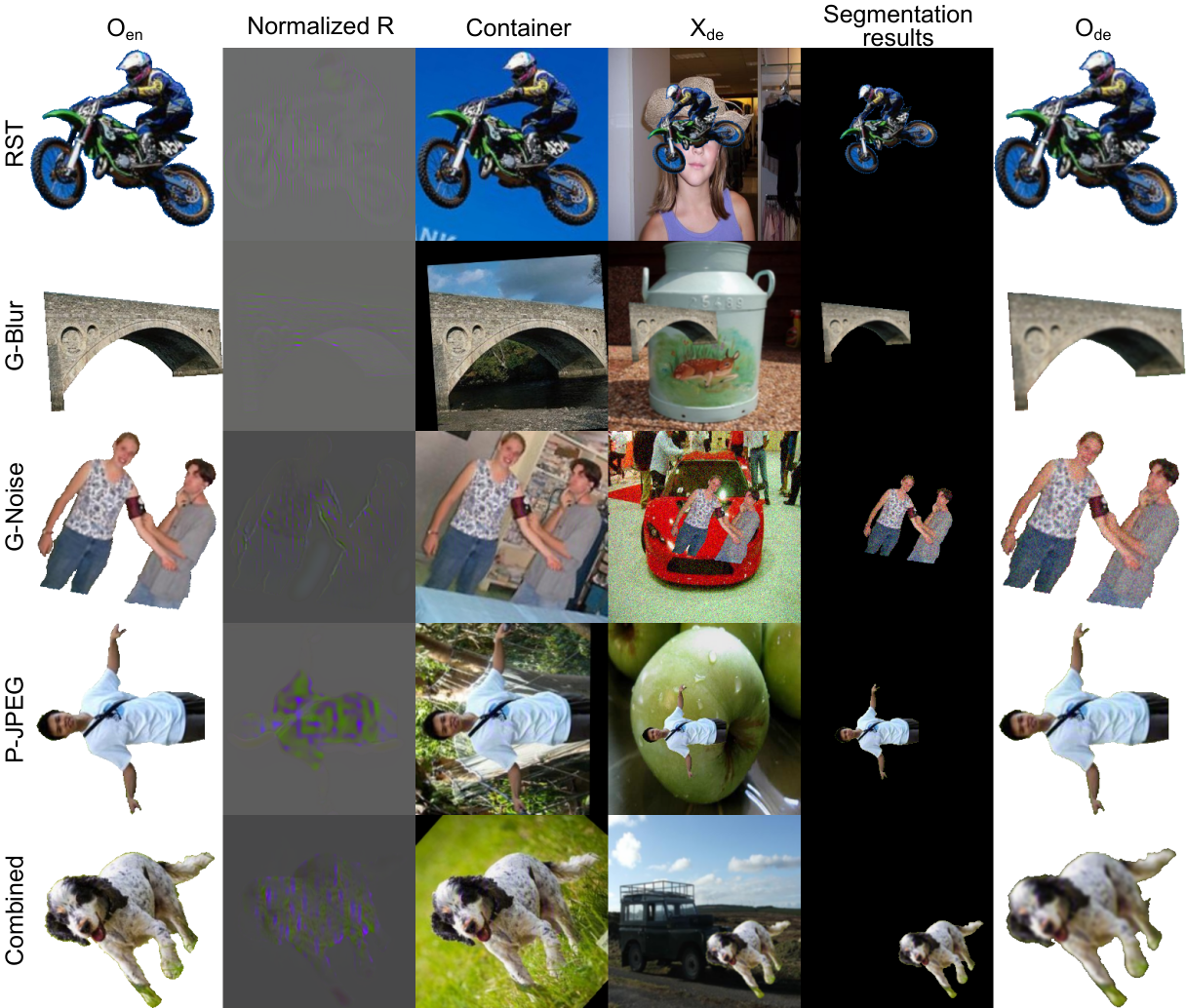}
	\centering
	\caption{Examples of the models trained with different noise layers. The container is the original image with the watermarked object. $O_{en}$ and $O_{de}$ should have a black background in training, we set them to white here for better visualization.}
	\label{fig:results} 
\end{figure}

\begin{table*}[t]
	\centering
	\caption{Comparison with SOTA methods. The distortions used here are: Gaussian Blur(sigma=3), Gaussian Noise(sigma=0.05), JPEG(QF $\in[10:10:90]$), Median Blur(kernel size=5), Salt-Pepper Noise(ratio=0.1), Brightness(scale $\in[0.8, 1.2]$), Contrast(scale $\in[0.8,1.2]$), Saturation(scale $\in[0.8,1.2]$), Hue(scale $\in[-0.1,0.1])$. When performing the cropping-paste attack for other methods, we only perform random rotation and scaling. As they are not trained for object cropping, the real cropping-paste attack causes 50\% BAR. The parameters are the same as ours. $\dag$: denotes the distortions used in the training process. *: the values are borrowed from the original manuscript.}
	\resizebox{\linewidth}{!}{
		\begin{tabular}{c|ccc|ccccccccc|c} \hline
			& \makecell[c]{PSNR\\(dB)} & \makecell[c]{SSIM\\(\%)}  & \makecell[c]{Capacity\\($\times 10^{-3}$ bpp)} & G-Blur & G-Noise & JPEG & M-Blur & S.P. & Bri. & Con. & Sat. & Hue & Crop-Paste  \\ \hline
			
			RoSteALS\cite{conf/cvpr/BuiAYC23} & 35.92 & 97.06 & 1.53 (100@256) & 98.04$^\dag$ & 97.72$^\dag$ & \textbf{95.25}$^\dag$ & 98.35 & 85.18 & 80.37$^\dag$ & 80.54$^\dag$ & 96.36$^\dag$ & \textbf{98.16}$^\dag$ & 51 \\
			ARWGAN\cite{journals/tim/HuangLLYXC23} & 39.07 & 98.33 & 1.83 (30@128) & 63.50 & 92.83$^\dag$ & 87.67$^\dag$ & 88.89 & 66.94 & 86.08 & 86.31 & 90.43 & 96.92 & 63.50 \\
			OBW*\cite{journals/mta/GajPS16} & 39.7 & - & - & $\approx$85 & $\approx$68 & $\approx$75 & - & - & - & - & - & - & $\approx$85 \\
			ARWGAN$_{rs}$ & 35.59 & 96.75 & 1.83 (30@128) & 84.64$^\dag$ & 98.07$^\dag$ & 87.42$^\dag$ & 95.01 & 73.89 & 82.55 & 82.69 & 87.91 & 93.04 & 95.98$^\dag$ \\
			SSyncOA$_{128}$ & \textbf{40.03} & \textbf{99.30} & 1.83 (30@128) & \textbf{98.61$^\dag$} & \textbf{98.59$^\dag$} & 93.54$^\dag$ & \textbf{98.75} & \textbf{98.29} & \textbf{96.95} & \textbf{98.84} & \textbf{98.90} & 96.20 & \textbf{98.70}$^\dag$ \\
			\hline
		\end{tabular}
	}
	\label{tab:comparison}
\end{table*}

\begin{figure}[t]
	\includegraphics[width=\linewidth]{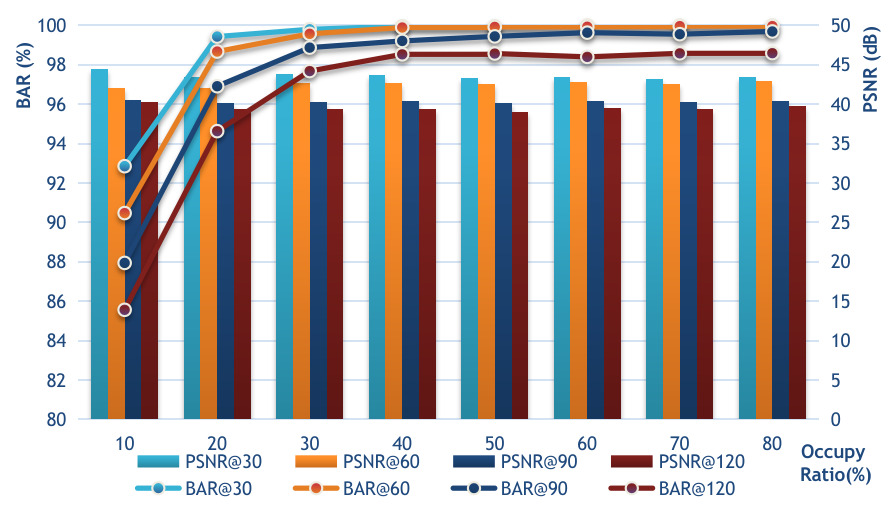}
	\centering
	\caption{Embedding capacity evaluation of SSyncOA. There are four watermarking models configured with capacity settings for embedding 30, 60, 90, and 120-bit messages within the objects in images of size 256.}
	\label{fig:capacity}
\end{figure}
\subsection{Visual quality,  Robustness and Capacity}
\textbf{Performance against different distortions.} We first train four distortion-specific watermarking models and a combined model. The distortion in each iteration of the combined model is randomly selected from the four noise layers. They are tested under the same conditions as their training process. The quantitative results are shown in Table \ref{tab:result}. It can be found that our method exhibits excellent visual quality, i.e., at least 43 dB in PSNR. This is primarily owed to our self-synchronization scheme, which eliminates the need for redundant information embedding.
In terms of robustness, JPEG compression has the largest impact, resulting in an error rate of about 3\%. We present some examples in Fig.\ref{fig:results}. It shows that our encoder has learned well to align the watermark with the object. As far as we know, this is the first CNN-based object watermarking model.

\textbf{The impact of segmentation results on decoding.} We show the test result of our watermark segmentation model as well as the decoding accuracy BAR$_{gt}$ tested with the ground truth segmentation masks in Table \ref{tab:result}. It can be seen that better segmentation result benefits the decoding accuracy. The small discrepancy between BAR and BAR$_{gt}$ also illustrates the robustness of our method against segmentation bias.

\textbf{The capacity of the object-aligned watermarking.} The capacity, i.e. \#bits/\#pixels (bits-per-pixel, bbp), depends on the message length and the object size. Here, we train three more watermarking models under the combined noise layer with the message length set to 60, 90, and 120, respectively.  During testing, we group the test dataset according to the size of the objects and calculate the BAR and PSNR under different image occupancy ratios. Results are presented in the Fig.\ref{fig:capacity}. It shows that bit accuracy increases with the object size when the message length is fixed, while visual quality is degraded due to more modifications. With the BAR of 98\%, our method can embed up to 120 bits of message in 19660 pixels (30\% of 256 $\times$ 256), achieving a bits-per-pixel of 6.10$\times$10$^{-3}$. 


\subsection{Compared with Others}
Two image watermarking methods, RoSteALS \cite{conf/cvpr/BuiAYC23} and ARWGAN \cite{journals/tim/HuangLLYXC23}, and a traditional object watermarking method, i.e. OBW \cite{journals/mta/GajPS16} are adopted here for comparison. 
RoSteASL proposes to embed watermarks in the image latent space to achieve robustness, and ARWGAN is a recent SOTA deep watermarking model embedding in the spatial domain. We re-evaluate their released models on our test dataset. 
For fair comparison, we retrained our model on image size $128 \times 128$, which is denoted as $\text{SSyncOA}_{128}$. 
We also retrained ARWGAN and RoSTeALS with the same noise layer settings as ours (excluding cropping and translation). But only the ARWGAN is converged, we call it ARWGAN$_{rs}$.


The comparison results are shown in Table \ref{tab:comparison}. Our model achieves significant progress compared to others in both overall robustness and visual quality. This is mainly because the proposed self-synchronization module allows our model to decode in a synchronized manner, and the object-aligned watermarking model also reduces the amount of pixel modification in images.  Results of ARWGAN$_{rs}$ also illustrate that improving the model's robustness against superimposed desynchronization distortions through training causes serious visual quality degradation. 

\begin{figure}[t]
	\includegraphics[width=0.9\linewidth]{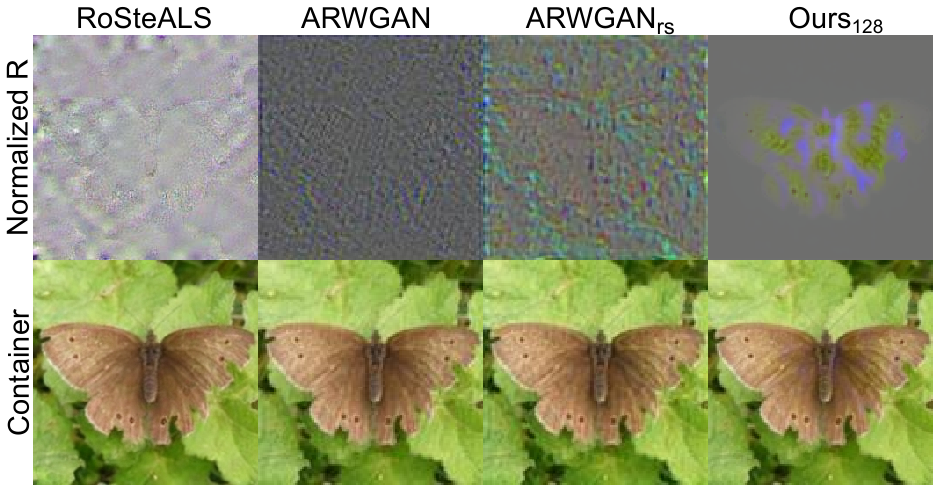}
	\centering
	\caption{Comparison of different methods in visual quality.}
	\label{fig:fig_compare}
\end{figure}

We present examples of normalized residual and container images of different methods in Fig.\ref{fig:fig_compare}. It can be seen that other methods need to embed redundant textures to resist distortions, particularly the desynchronization distortion, whereas our method modifies only a few pixels in the object region.


\subsection{Ablation of synchronizations}
Ablation experiments are performed here to verify the benefit of the four types of synchronizations on watermark training. Based on SSyncOA, for ID-1, we conduct the cropping synchronization with a probability of 0.5 during training. This allows the model to embed watermarks in the entire image while having the ability to resist cropping; 
for ID-2/3/4, we remove the rotation/scaling/translation synchronization respectively. 
For ID-5, without any geometric synchronization, we simply crop out the object as the encoder and decoder inputs during training. To clarify their performance discrepancy, the message length is set to 60 here.

The results are shown in Table 3. It indicates that desynchronizing rotation and scaling significantly degrades the performance of the watermarking model, specifically, the PSNR drops by almost 7 dB, and the BAR is also degraded. Conversely, the impact of cropping and translation is relatively small, potentially attributed to CNN's translation invariance. Even when fed with only objects without any synchronization, the model still converges, but the performance of the trained model experiences a sharp decline compared to that with synchronization.



\begin{table}[t]
	\centering
	\caption{Ablation study for different synchronizations.}
	\resizebox{0.95\linewidth}{!}{
		\begin{tabular}{c|cccc|ccc} \hline
			ID & \makecell[c]{Sync\\Cropping} & \makecell[c]{Sync\\Rotation} & \makecell[c]{Sync\\Scale} & \makecell[c]{Sync\\Translation} & PSNR & SSIM & BAR$_{gt}$ \\ \hline
			1 & \XSolidBrush & & &  & 41.46 & 99.07 &  97.55\\
			2 & & \XSolidBrush &  &  & 36.85 & 98.57 &  95.55\\
			3 & & & \XSolidBrush &  & 35.47 & 98.91 & 96.44 \\
			4 & & &  & \XSolidBrush & 39.13 & 99.23 & 97.55  \\
			5 & & \XSolidBrush & \XSolidBrush & \XSolidBrush & 34.64 & 96.24 & 94.37 \\
			6 & & &  &  & \textbf{42.38} & \textbf{99.36} & \textbf{97.90}\\
			\hline
		\end{tabular}
	}
	\label{tab:ablation}
\end{table}
\section{Conclusion}
\label{sec:conclusion}
In this paper, we propose a self-synchronized object-aligned watermarking scheme, called SSyncOA, designed to protect the object copyright against cropping-paste attacks. To remove superimposed desynchronization distortions, we align the watermark region with the cropped object and achieve geometric synchronization by normalizing the object's invariant features during both encoding and decoding processes. The self-synchronization process and the cropping-paste attack are integrated into the end-to-end watermark training process, enabling our object-aligned model to embed and extract watermarks based on the object region. Extensive experiments demonstrate the superiority of SSyncOA over other SOTAs.



\footnotesize
\bibliographystyle{IEEEbib}
\bibliography{icme2023template}

\end{document}